\documentclass[sigconf]{acmart}
\acmSubmissionID{1127}

\usepackage[ruled]{algorithm2e} 
\usepackage{xcolor}
\usepackage{stfloats}
\usepackage{optidef}
\usepackage{multicol}
\usepackage{float}
\usepackage{dsfont}

\AtBeginDocument{%
  }

\copyrightyear{2024}
\acmYear{2024}
\setcopyright{rightsretained}
\acmConference[SIGGRAPH Conference Papers '24]{Special Interest Group on Computer Graphics and Interactive Techniques Conference Conference Papers '24}{July 27-August 1, 2024}{Denver, CO, USA}
\acmBooktitle{Special Interest Group on Computer Graphics and Interactive Techniques Conference Conference Papers '24 (SIGGRAPH Conference Papers '24), July 27-August 1, 2024, Denver, CO, USA}
\acmDOI{10.1145/3641519.3657517}
\acmISBN{979-8-4007-0525-0/24/07}

\begin{document}

\title{Physics-based Scene Layout Generation from Human Motion}

\author{Jianan Li}
\orcid{0009-0009-5684-9381}
\thanks{The work was done while Jianan Li was an intern at Tencent.}
\email{jnli22@cse.cuhk.edu.hk}
\affiliation{%
  \institution{The Chinese University of Hong Kong}
  \city{Hong Kong}
  \country{China}
}
\affiliation{%
  \institution{Tencent Robotics X}
  \city{Shenzhen}
  \country{China}
}

\author{Tao Huang}
\orcid{0009-0006-7403-2760}
\email{thuang22@cse.cuhk.edu.hk}
\affiliation{%
  \institution{The Chinese University of Hong Kong}
  \city{Hong Kong}
  \country{China}
}

\author{Qingxu Zhu}
\orcid{0009-0001-6549-3438}
\email{qingxuzhu@tencent.com}
\affiliation{%
  \institution{Tencent Robotics X}
  \city{Shenzhen}
  \country{China}
}

\author{Tien-Tsin Wong}
\orcid{0000-0002-7792-9307}
\email{ttwong@cse.cuhk.edu.hk}
\affiliation{%
  \institution{The Chinese University of Hong Kong}
  \city{Hong Kong}
  \country{China}
}

\renewcommand{\shortauthors}{Jianan Li, Tao Huang, Qingxu Zhu and Tien-Tsin Wong}

\begin{abstract}
Creating scenes for captured motions that achieve realistic human-scene interaction is crucial for 3D animation in movies or video games. 
As character motion is often captured in a blue-screened studio without real furniture or objects in place, there may be a discrepancy between the planned motion and the captured one. 
This gives rise to the need for automatic scene layout generation to relieve the burdens of selecting and positioning furniture and objects. Previous approaches cannot avoid artifacts like penetration and floating due to the lack of physical constraints. Furthermore, some heavily rely on specific data to learn the contact affordances, restricting the generalization ability to different motions.
In this work, we present a physics-based approach that simultaneously optimizes a scene layout generator and simulates a moving human in a physics simulator. To attain plausible and realistic interaction motions, our method explicitly introduces physical constraints. To automatically recover and generate the scene layout, we minimize the motion tracking errors to identify the objects that can afford interaction. We use reinforcement learning to perform a dual-optimization of both the character motion imitation controller and the scene layout generator. To facilitate the optimization, we reshape the tracking rewards and devise pose prior guidance obtained from our estimated pseudo-contact labels. We evaluate our method using motions from SAMP and PROX, and demonstrate physically plausible scene layout reconstruction compared with the previous kinematics-based method.
\end{abstract}


\begin{teaserfigure}
    \centering
    \includegraphics[width=\textwidth]{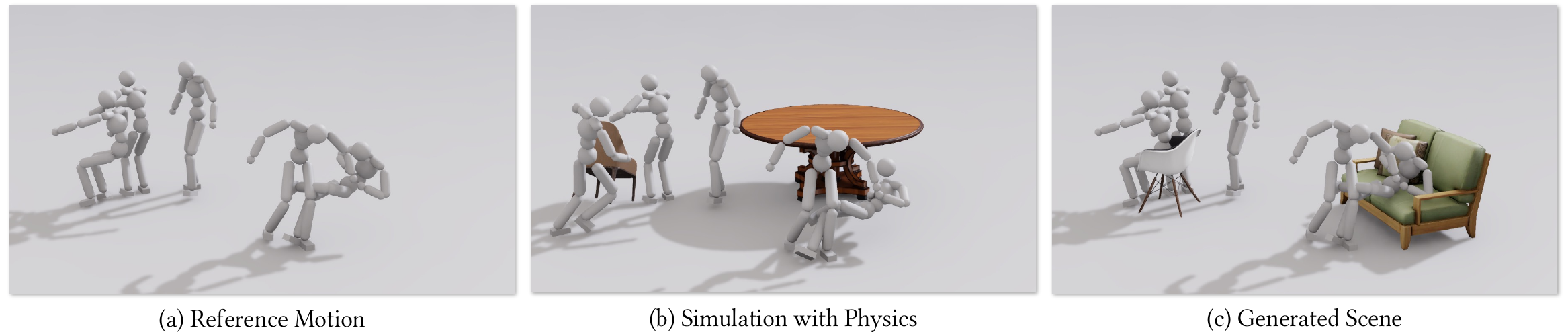}
    \caption{\textbf{(a)} Given a captured 3D human motion as the reference, \textbf{(b)} our proposed framework optimizes the scene configuration to reproduce the physical interactions in simulation, and then generates suitable affording objects, \textbf{(c)} allowing physically plausible interaction for the animated character in the scene.}
    \label{fig:teaser}
\end{teaserfigure}

\maketitle

\section{Introduction}

Even with the advanced motion capture facilities, creating realistic animated avatars that seamlessly interact with surrounding objects poses a significant challenge for animators involved in film production and game development. Typically, character motion is blue-screen captured in a studio without the actual physical furniture or objects in place. Even if the virtual furniture/objects have been planned ahead, there could be deviations during the actual capture. In other words, an animator may still have to meticulously choose a suitable piece of furniture (e.g. a chair), position it correctly, and adjust the human motion to fit it. Hence, it would be convenient to automate this tedious procedure of generating a plausible scene for a given captured human motion thereby achieving natural human-scene interaction. 

Several studies have explored the synthesis of scene layouts from human motions. \citet{nie2022pose2room} proposed a data-driven approach that learns a probabilistic distribution of room layout conditioned by human skeleton pose trajectory. However, this method can only output semantic bounding boxes of the room layout without modeling the contact relationship between the human and objects. To generate a visually compelling scene with a moving human, mesh-based human models and post-optimization for object placement have been used in \cite{yi2022human, ye2022scene, yi2023mime}. \citet{ye2022scene} presented a two-stage pipeline that initially estimates contact vertices on human bodies and subsequently recovers the affording object by minimizing contact and collision losses based on the object and the human meshes. However, such soft constraints cannot ensure a physically correct interaction. Physics violations, such as interpenetration and floating, may still occur, particularly when the character comes into contact with the object. Moreover, the reliance on learned contact semantics predictor might restrict its applicability to specific types of motions, and the diversity of object categories within the synthesized scene is also limited.

\begin{figure}[t]
    \centering
    \includegraphics[width=0.47\textwidth]{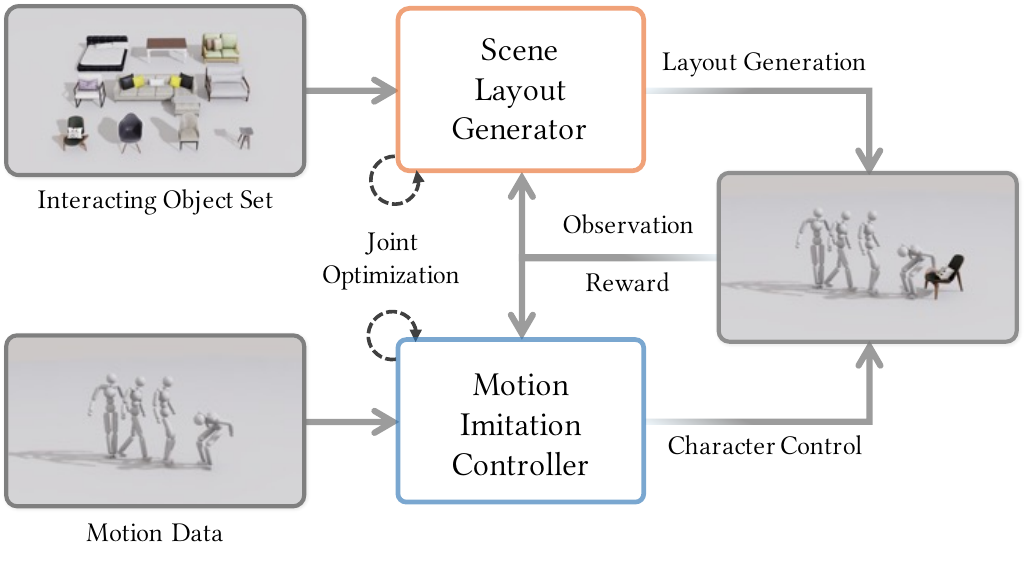}
    \caption{Given a human motion sequence and an interacting object set, our framework performs joint optimization for the two composed parts: a motion imitation controller that controls the simulated character and a scene layout generator that configures the objects in the simulation environment.}
    \label{fig:overview}
\end{figure}

To obtain physically plausible interactions within the generated scene, we propose to impose stricter physics constraints by simulating a virtual character in a physics-based environment. In addition, to generate a reasonable scene for arbitrary motions with any possible affording objects, we propose to infer the contacting object based on the physical relations between the human and the scene. The underlying intuition is that the human needs the object to support the interaction in the real physical world. In this paper, we focus on generating the interacting objects, which is the most critical part of the scene that a human is interacting with. We present Simultaneously Inferring the Interacting Objects and Learning Human-Scene Interaction Motions (INFERACT), a physics-based optimization framework, that holistically synthesizes the scene layout and imitates the motion within a physics simulator. As illustrated in Fig.~\ref{fig:overview}, our framework consists of two modules, a motion imitator which learns a character controller to imitate the input motion, and a scene layout generator that predicts suitable contacting objects with their placements. We perform dual-optimization using reinforcement learning and adopt the motion tracking reward as the objective for both sides. Finding the optimal object placement is challenging due to the inefficient random exploration strategy of the reinforcement learning agent. To mitigate this issue, we first incorporate the contact constraint into the tracking reward to encourage frequent interactions between the human and the object. Then, we leverage the contact human poses as pose priors to further provide stronger guidance for the object placement, by estimating contact frames using an implicit function of contact frame values.

We evaluated the efficacy and effectiveness of the proposed method on motions captured in the PROX, SAMP datasets, and other outdoor motions. Our experiments demonstrate that the proposed method excels in generating physically plausible scene layouts that align with the given motion. Additionally, we show that our approach is capable of generating diverse scene configurations and selecting appropriate objects for various motions. We demonstrate the direct applicability of our approach to other interaction motions, such as vaulting, without the need for additional supporting data. In the ablation study, we rigorously verified the effectiveness of the character-object contact constraint and the pose prior guidance.

\section{Related Works}
\paragraph{Human-Scene Interaction Synthesis}
Research on human-scene interactions (HSI) seeks to precisely model the relationship between a human character and the interacting object. One strand of this research focuses on synthesizing plausible static human poses within a given scene~\cite{grabner2011makes, gupta20113d, savva2016pigraphs, hassan2019resolving, li2019putting}. To enhance the realism of the generated human figures, considerations are given to salient interactions such as contact and proximity between the human and the scene. \citet{zhang2020place} proposed an HSI synthesis framework that explicitly incorporates human-scene contacts into the pose generation procedure, yielding realistic interactions. POSA~\cite{hassan2021populating} introduces a body-centric HSI representation that encodes semantic contacts onto the mesh vertices of a parametric human body model, SMPL-X~\cite{SMPL-X:2019}.

Another line of work focuses on modeling the dynamic process of human-scene interactions for realistic animated interactions. \citet{starke2019neural} proposed a goal-conditioned motion synthesis model that achieves precise scene interactions and motion control. \citet{hassan2021stochastic} presented a stochastic model enabling the generation of diverse styles of sitting and lying motions. COUCH~\cite{zhang2022couch} employs a contact-conditioned variational autoencoder, facilitating fine-grained control over the interaction motion between the human and the chair. Canonical motions are highlighted in \cite{mir2023generating}, which enables the generation of continuous human-scene interaction motions using scene-agnostic MoCap data. In recent works, diffusion models have been applied to capture HSI patterns, offering enhanced motion quality and flexible editing possibilities~\cite{huang2023diffusion, kulkarni2023nifty, ye2023diffusion, taheri2024grip}. Physics-based methods are also employed for human-scene interaction synthesis, achieving realistic and physically plausible interactions~\cite{hassan2023synthesizing, pan2023synthesizing}. In contrast to these endeavors, our work focuses on the generation of plausible scene layouts aligned with a given motion, resulting in life-like human-scene interactions.

\paragraph{Physics-based Motion Tracking}
Physics-based motion imitation is extensively employed for learning and reproducing human movements in physics-based environments through reinforcement learning~\cite{coros2009robust, peng2016terrain, peng2021amp, peng2022ase, wang2020unicon}. One straightforward approach to imitating a reference motion is motion tracking, wherein the primary objective is to minimize the pose error between the simulated character and a given reference~\cite{sok2007simulating, muico2011composite, liu2010sampling, liu2016guided, fussell2021supertrack}. Motion tracking has demonstrated its efficacy in acquiring fundamental motor skills, even those involving highly dynamic movements~\cite{peng2018deepmimic, bergamin2019drecon, won2020scalable, lee2021learning}. It also serves as an effective method for learning primitive skills in hierarchical models~\cite{merel2020catch, yao2022controlvae, won2022physics, zhu2023neural}. During the pre-training of latent hierarchical models, motion tracking rewards function akin to the reconstruction loss in variational autoencoders. Beyond its application in learning motion synthesis models, physics-based motion imitation is applied as a post-processing step to ensure physical plausibility~\cite{shimada2020physcap, shimada2021neural, xie2021physics, yuan2023physdiff}. In our approach, the utilization of the tracking-based character controller extends beyond refining motion artifacts and ensuring physically correct scene interactions. It also encompasses the generation of plausible scenes wherein the simulated characters engage in meaningful physical interactions.

\paragraph{Human-guided Scene Layout Generation}
Humans play a crucial role in guiding and facilitating the generation and reconstruction of indoor scene layouts. For scene layout reconstruction, ~\citet{chen2019holistic++} leveraged the inherent connection between the estimations of human poses and the bounding boxes of objects in a scene. They proposed merging these two individual tasks to enhance the scene understanding from videos. iMAPPER provides a pipeline based on motion retrieval that can produce realistic object layouts for videos even with severe occlusion~\cite{monszpart2019imapper}.~\citet{weng2021holistic} introduced a model that predicts mesh-level estimations for both humans and the scene, followed by joint optimization to refine the results. To achieve improved 3D scene layout reconstruction, ~\citet{yi2022human} explicitly incorporated Human-Scene Interaction (HSI) constraints into the optimization of object placement.
  
Concerning scene layout generation, ~\citet{nie2022pose2room} introduced an end-to-end generative model that takes human motion as inputs and predicts the bounding boxes of furniture in a room. A recent work~\cite{ye2022scene} achieved scene synthesis with object meshes by proposing a framework SUMMON that recovers the interacting object based on the contact semantics of the humans. MIME~\cite{yi2023mime} also leverages human contacts as conditional inputs of a transformer-based model to predict a room layout represented by a sequence of objects. To mitigate artifacts like penetration and floating, such kinematics-based methods often need an additional refinement stage mostly to minimize contact or collision losses, which might be insufficient to obtain physically correct results. In contrast, our proposed approach recovers and generates scene configurations within a physics-based environment to ensure physical plausibility. In addition, the physics constraints are unitized to steer the scene layout generation procedure in our devised dual-optimization framework, where reinforcement learning is used to explore the optimal object placement for interacting objects.
\begin{figure*}
    \centering
    \includegraphics[width=0.8\textwidth]{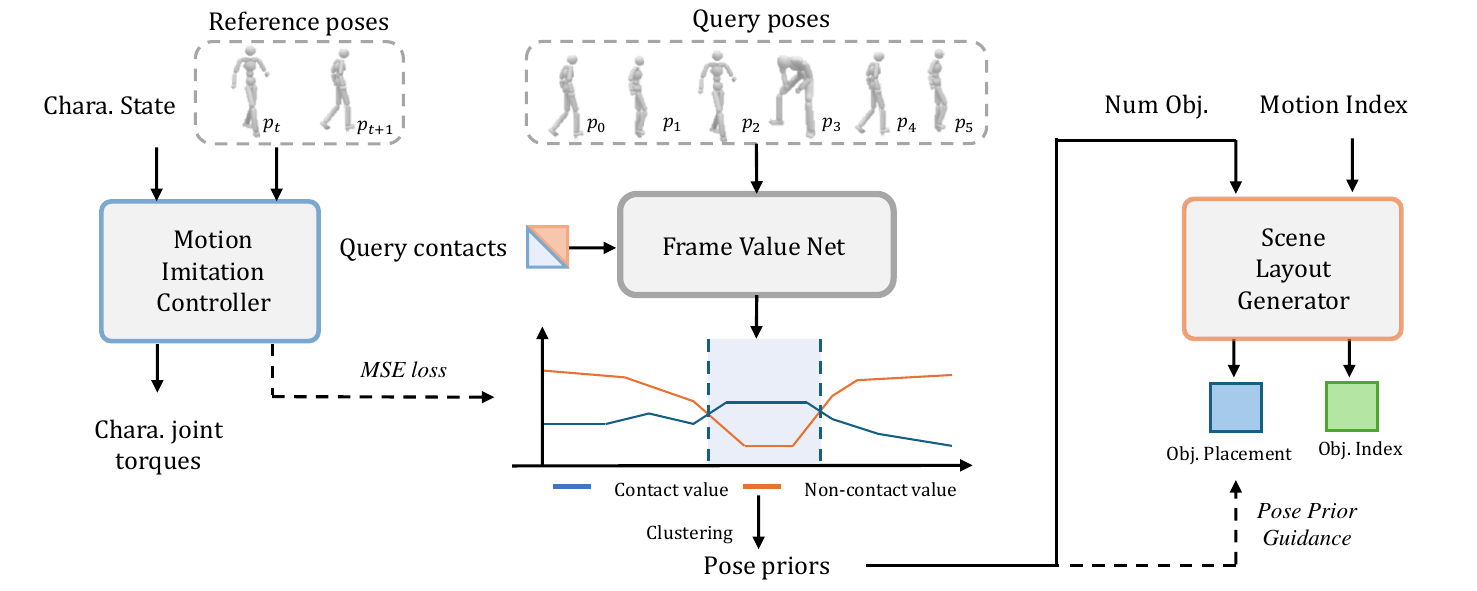}
    \caption{The motion imitation controller is trained to track the reference motion within a physics simulator. The frame value network, guided by the motion imitation controller, predicts the frame value for a given query pose and contact condition. Pseudo contact labels are derived by identifying frames with higher estimated contact frame values compared to non-contact frame values. Subsequently, pose priors and the number of contacting objects can be obtained through clustering of the contact human poses. The scene layout generator policy takes the motion index and the number of objects as inputs and predicts a mixed distribution of selected object indices and their corresponding placements, with the guidance of pose priors.}
    \label{fig:structure}
\end{figure*}

\section{Method}
Given a set of interacting objects and a motion sequence, we aim to generate scene layouts with physically plausible object placement that aligns with human actions. At a high level, our proposed framework formulates scene layout generation as an optimization problem for maximizing the motion tracking score in physics-based simulation. The optimization process comprises two key components: learning a motion imitation controller to animate the simulated character within a physics simulator and updating the scene layout generator to provide the physical affordances for the interaction. This section is structured as follows. Firstly, we introduce the problem formulation of scene-character joint optimization in Section~\ref{scene_character_optim}. Then, we present the details of the motion imitation controller in Section~\ref{motion_imitator}. Lastly, we introduce the scene layout generator with the character-object contact constraint and pose prior guidance for efficient scene layout optimization in Section~\ref{scene_layout}.

\subsection{Scene-Character Joint Optimization}\label{scene_character_optim}
To generate a scene that satisfies the interaction and contact relationship with the given moving human, we formulate the scene layout generation from human motions as a maximization of the motion tracking score with physics constraints. Given a human kinematics motion sequence with a length of $T$ timesteps $m=p_{1:T}$, a set of $N$ objects $\mathcal{O}=\{o_i\}_{i=1}^N$, we denote the actual motion of the simulated human character in a physics environment as $\hat{p}_{1:T}$ and represent the generated scene with $L$ interacting objects using a subset of given objects $\{o_{i_j}\}_{j=1:L}$ and their placements $\{q_j\}_{j=1:L}$. Our objective is to generate an appropriate scene for the given motion, in other words, finding an optimal selection of objects $\{o^*_{i_j}\}_{j=1:L}$ associated with the optimal locations $q^*(o^*_{i_j})_{j=1:L}$ that can support the animated character perfectly reproducing the reference motion $m$ under the physics constraints. To animate the simulated character within the physics simulator, we employ a motion imitation controller that minimizes the discrepancy between $\hat{p}_{1:T}$ and $p_{1:T}$. We thus propose a dual optimization framework to jointly optimize the motion imitation controller and the scene layout generator.

\subsection{Motion Imitation Controller}\label{motion_imitator}
To effectively reproduce the input kinematics motion on a physical human character, we train a motion imitation controller to perform physics-based motion tracking. Motion tracking in a physics environment can be formulated as a Markov Decision Process (MDP)~\cite{kaelbling1996reinforcement}, which can be solved using reinforcement learning. We represent the character controller as a $\pi_C(s_t)$, which outputs control commands $a_t$ as actions based on the observation of the current state $s_t$. The physics simulator calculates the subsequent state $s_{t+1}$ based on the applied action and the environment dynamics $p(s_{t+1}|s_t, a_t)$. Additionally, it provides a reward $r_t$ to the character controller. By repeating these steps, a trajectory of observed states $\tau = \{s_0, s_1, …, s_{T-1}\}$ is collected by the policy. The policy is then updated to maximize the cumulative discounted rewards $R=\sum_{t=0}^{T-1} \gamma^tr_t$.
\paragraph{State and Action Representation} In physics-based motion tracking, we represent the state observation of the policy with three components $s_t=[s_t^H, s_t^E, s_t^R]$, where $s_t^H \in \mathbb{R}^{125}$ is a local proprioceptive observation of the human character's pose, $s_t^E \in \mathbb{Z^+} \times \mathbb{R}^{3}$ represents the state of a scene layout and $s_t^R \in R^{127}$ denotes the relative tracking feature of the reference motion. More specifically, the scene layout state consisting of $L$ objects is a sequence of object indices $i_j \in \mathbb{Z}^+$ and their placements $q_{i_j} \in \mathbb{R}^3$ for each object $j$. The details of computation for each state component can be referred to in the appendix. We use a commonly used humanoid character~\cite{peng2022ase} with 12 active joints and a total of 28 degrees of freedom (DoFs). To control the character in the simulation, we calculate the driving torques for each DoF using a PD controller, whose input target joint position is defined as the action $a_t \in \mathbb{R}^{28}$.
\paragraph{Rewards} We adopt a similar reward function presented in \cite{peng2018deepmimic} as the motion tracking reward, which is given by
\begin{equation}\label{eq:tracking_rew}
    r_t = w^pr^p_t + w^or^o_t + w^vr^v_t + w^{jp}r^{jp}_t + w^{jv}r^{jv}_t + w^kr^k_t,
\end{equation}
where $r^p_t$, $r^o_t$, $r^v_t$ are the character's root position, root orientation, and root velocity, and $r^{jp}_t$, $r^{jv}_t$, $r^k_t$ denote the joint positions, joint velocities, and positions of key bodies on the character respectively.

\subsection{Scene Layout Generator}\label{scene_layout}
To populate plausible scenes for the simulated character to interact, we optimize a scene layout generator simultaneously while learning the motion imitation controller. We represent the scene layout generator with a stochastic policy that outputs a joint distribution of discrete object indices and continuous object placements. The input to the scene layout generator can be either a complete motion sequence or other hand-crafted features that represent the reference motion. For simplicity, we use a motion index $I(m) \in \mathbb{Z^+}$ as a dummy input in our implementation. For a given motion index $I(m)$, the scene layout generator predicts the selection and the placement of the $j$-th object in contact with the human character as $(i_j, q_{i_j}) \sim \pi_S(I(m), j)$, where index $i_j$ indicates the object selection from the object set $\mathcal{O}$.

To train the scene layout generator, we use reinforcement learning to maximize the accumulated motion tracking rewards, similar to the learning of the motion imitation controller. However, tracking rewards cannot provide dense feedback to the reinforcement learning agent, especially when the simulated character does not have actual physical interaction with the object. 

To facilitate learning, we incorporate guidance at different levels into reward functions to provide effective feedback to the scene layout generator. Firstly, we integrate a contact constraint into the motion tracking rewards, ensuring a close spatial relationship between the character and the interacting object. Secondly, we introduce direct guidance for object placement by utilizing pose priors derived from the contact human poses, estimated through an unsupervised contact frame value estimator.

\paragraph{Character-object Contacting Constraint}
To promote increased interaction between the character and the scene, we modify the reward structure for the scene layout generator by introducing a binary multiplier to the original motion tracking reward.
More specifically, we provide rewards to the scene generator only when the simulated human makes contact with the objects in the generated scene. The reshaped reward is represented by the equation:

\begin{equation}\label{eq:contact_constraint}
    R_{\rm track}=\mathds{1}(\sum_{t=1}^T c_t > 0)\sum_{t=1}^T \gamma^t r_t,
\end{equation}
where $c_t \in \{0, 1\}$ denotes the contact state between the human and the object at the time step $t$ and $\mathds{1}(\sum_{t=1}^T c_t > 0)=1$ if the internal statement is true, otherwise takes zero. With this reward shaping, the original motion tracking task is reformulated to a constrained tracking score maximization, where the primary requirement is to ensure the scene-character contact.

\paragraph{Pseudo Contact Labels}
The introduced character-object contacting constraint serves as a coarse prior regarding the spatial distribution of the human's location. To obtain a more precise object location from the motion, we can leverage frame-wise contact information that provides insights into the position of the interacting human. In contrast to previous approaches that rely on a trained contact estimation model~\cite{ye2022scene, yi2023mime}, we propose an unsupervised method to infer the contact pose based on the character's tracking performance. We specifically train an implicit contact frame value function, denoted as $V(p_t, c_t) \in \mathbb{R}$, to predict the expected tracking performance when provided with given human pose query $p_t$ and contact query $c_t$. To train the implicit contact frame value function, we utilize the estimated tracking performance value from the critic network~\cite{schulman2017proximal} to provide supervision signals. The training objective for the frame value network is presented as follows:
\begin{equation}
    \mathcal{L}_{\rm frameval} = \|V_{\phi}(p_t, c_t) - r_t - \gamma \hat{V}(s_{t+1})\|_2^2,
\end{equation}
where $V_{\phi}(p_t, c_t)$ represents the frame value network, $r_t$ denotes the tracking reward, and $\hat{V}(s_{t+1})$ indicates the estimated value for successive state $s_{t+1}$, obtained from the critic network of the motion imitation controller.

We utilize the frame value network to infer pseudo contact labels for the given motion sequence. As illustrated in Fig.~\ref{fig:structure}, by providing the pose and contact queries as inputs to the frame value network, we obtain two curves representing estimated frame values with and without contact. Subsequently, pseudo contact labels $\hat{c}_t$ can be computed by comparing the predicted contact frame values using the following equation:
\begin{equation}
    \hat{c}_t = \mathds{1}\big(V_{\phi}(p_t, 1) - V_{\phi}(p_t, 0) > 0\big).
\end{equation}

\paragraph{Contact Pose Priors}
With the estimated pseudo contact labels, we can incorporate stronger guidance for object placement by leveraging contact pose priors, to enhance the learning efficiency of the scene layout generator. The pose prior is represented by the pelvis pose of the contact humans~\cite{ye2022scene}, which can provide a rough indication of the object's location during the interaction. Besides, incorporating pose priors can enhance the quality of the solution in cases where the motion tracking objective alone may not be sufficient to achieve realistic outcomes. 

To derive pose priors, we perform clustering on the pelvis poses of the contact humans 
identified through the estimated pseudo contact labels. The pose priors are represented as cluster centers, denoted as $\bar{q_r}=(t_x, t_y, r_{\text{yaw}})$, which consists of planar translations and a vertical rotation angle. Moreover, the number of clusters can be used as an indication of the number of interacting objects present in the scene. We combine the reshaped tracking reward in Eq.~\ref{eq:contact_constraint} with the pose prior guidance to obtain the final reward function for the scene layout generator, which is shown as below:
\begin{equation}
    R = R_{\text{track}} + \alpha \|q - \bar{q_r}\|_2^2,
\end{equation}
where the weight $\alpha$ controls the strength of the pose prior guidance.

\begin{figure*}[t]
    \centering
    \includegraphics[width=\textwidth]{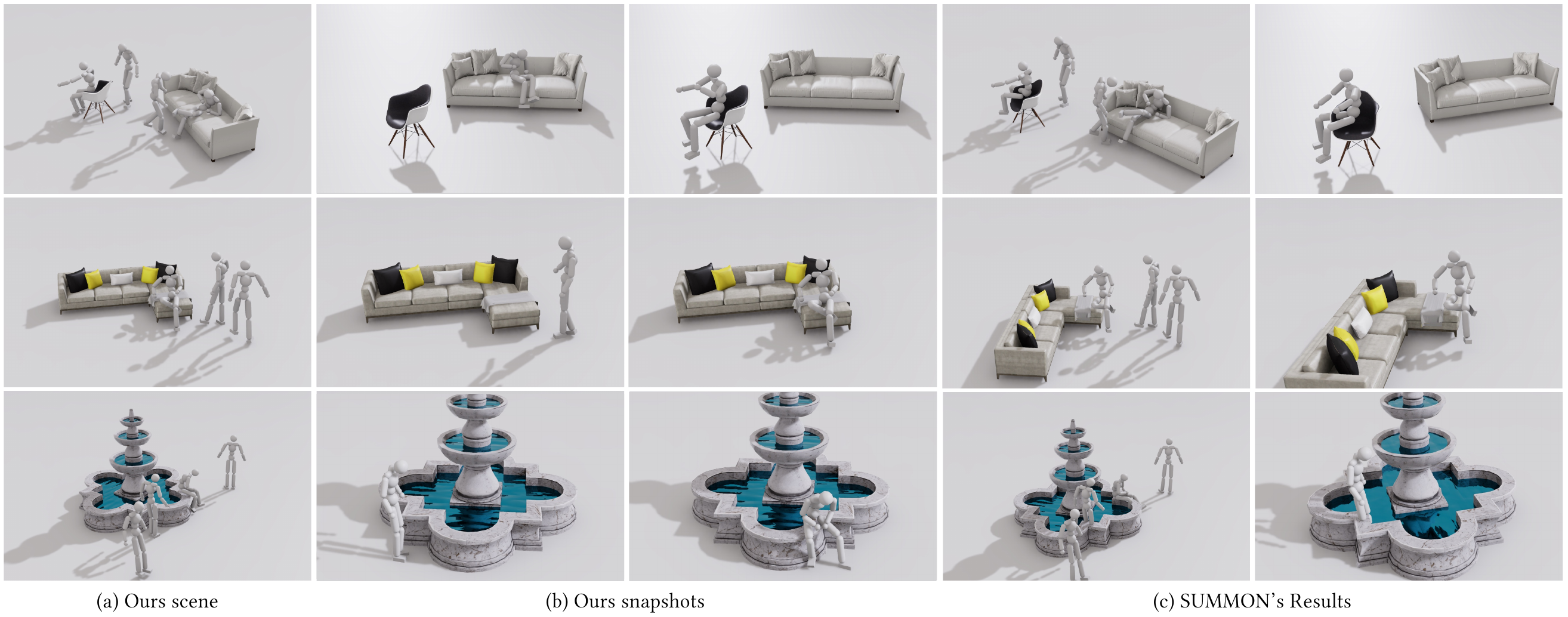}
    \caption{Illustrations of our generated scenes for three different motions, with the object placements obtained by SUMMON~\cite{ye2022scene} for comparison. Our method generates reasonable and plausible scenes compared with SUMMON.}
    \label{fig:comparison}
\end{figure*}

\begin{figure*}[t]
    \centering
    \includegraphics[width=\textwidth]{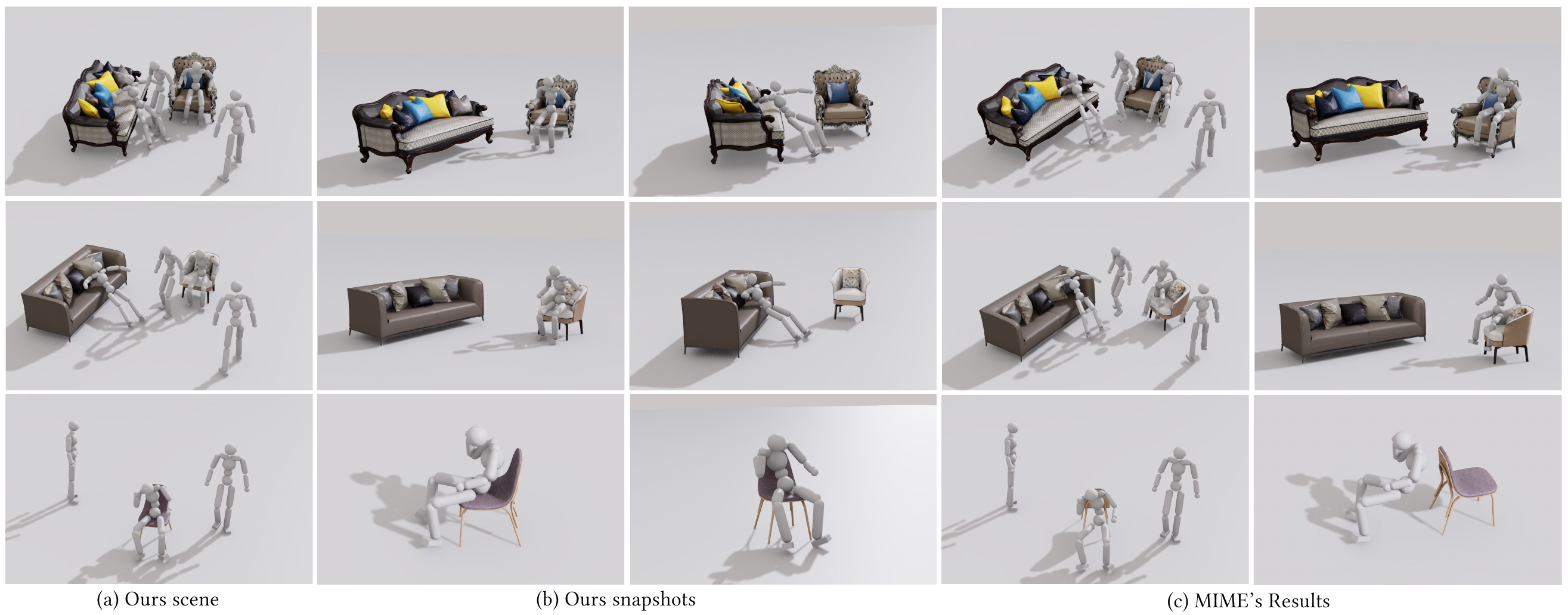}
    \caption{Qualitative comparison with MIME~\cite{yi2023mime}. The results exclusively focus on the contacting objects within the scene. Our method demonstrates enhanced physical plausibility in object placements, while MIME occasionally encounters difficulties in generating scenes with satisfactory human-scene interaction, even after employing scene refinement.}
    \label{fig:comparison_mime}
\end{figure*}

\begin{figure*}[t]
    \centering
    \includegraphics[width=\textwidth]{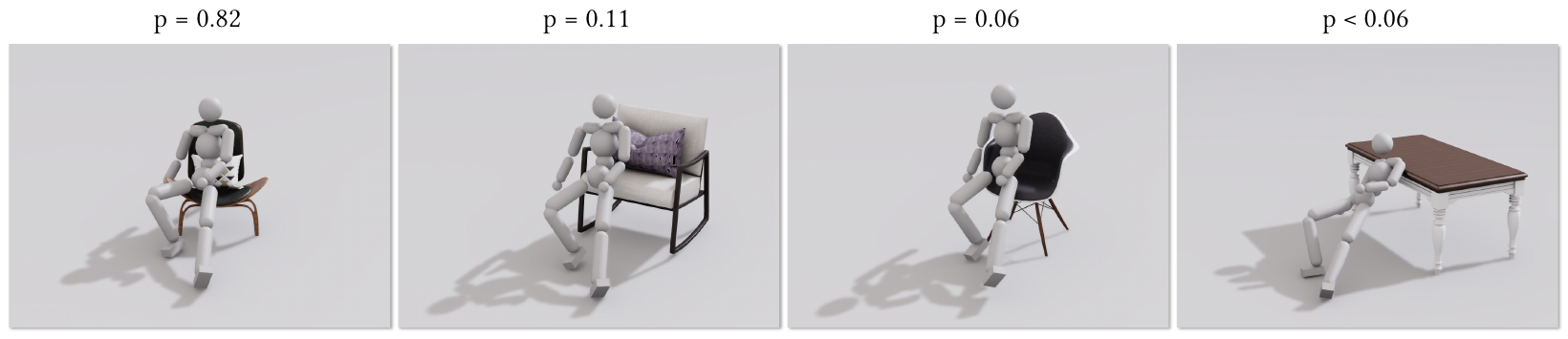}
    \caption{Visualizations of diverse object selections generated by our method for a sitting motion. The selection probability of each chosen object is indicated above the figures. This example demonstrates the capability of INFERACT to generate diverse results for chair choices (sub-figure 1 to 3) and its ability to screen out inappropriate objects like tables (rightmost).}
    \label{fig:object_selection}
\end{figure*}

\begin{figure*}[t]
    \centering
    \includegraphics[width=\textwidth]{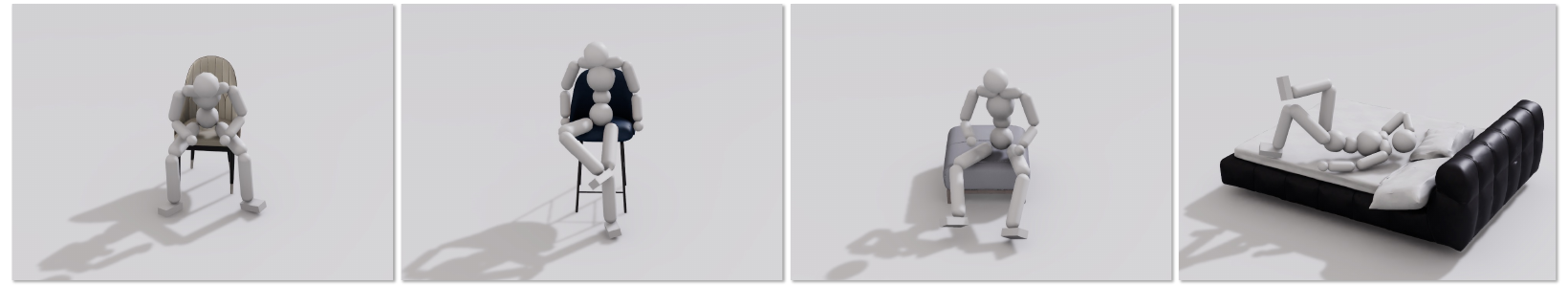}
    \caption{Illustrations of best selections for various motions. INFERACT can select the suitable affording object for different motions. Motions from left to right are sitting on a chair, sitting on a high stool, sitting on a footstool, and lying on a bed.}
    \label{fig:objects}
\end{figure*}
\section{Experiments}

\subsection{Experiment Setup}
We evaluate our approach using indoor motions from SAMP~\cite{hassan2021stochastic} and PROX~\cite{hassan2019resolving}, as well as outdoor vaulting motion capture data. Meanwhile, we prepare an object set consisting of 32 different furniture items from the 3D-Future dataset~\cite{fu20213d}, categorized into chairs, sofas, tables, and beds. Furthermore, we test other 3D shapes such as fountains and rocks to evaluate the ability of our method to generate reasonable placements, even when explicit affordances for interaction may not be present.

\subsection{Implementation Details}
The motion imitation controller $\pi$ is represented by a neural network with three fully connected layers of units [512, 256, 64]. The output of the policy $\pi$ is a Gaussian distribution $\pi(s_t) = \mathcal{N}(\mu(s_t), \Sigma)$, where the mean $\mu(s_t)$ is predicted by the neural network, and the variance $\Sigma$ is a constant diagonal matrix defined manually. To handle the scene layout state inputs $s_t^E$, which consists of an index input $i_j$ and a placement input $q_{i_j}$ for each object in the scene, we project $s_t^E$ into 128D vectors using an embedding layer and a linear layer. Similarly, the scene layout generator is also represented by a neural network. It consists of two fully connected neural networks with sizes [64, 16]. The first network takes inputs of the motion index $I(m_i)$ and the object order $j$, and then outputs a categorical distribution that represents the selection probability for each object in the set $\mathcal{O}$. Once an object is chosen, the second neural network predicts the corresponding placement. The frame value network $V{\phi}(p_t, c_t)$ is modeled using fully connected layers of [256, 128, 64]. The input pose query $p_t$ is a feature of a canonical pose from the reference motion, where the pelvis translation is set to zero to remove global information.

The optimization of the whole proposed framework is based on reinforcement learning. We create a training environment in a GPU-based parallel simulator IsaacGym~\cite{makoviychuk2021isaac}. The contacts between the character and objects are detected using the force sensor integrated into the simulator. To accelerate the learning process of the motion imitation controller, we adopt a reference-based initialization strategy recommended in~\cite{peng2018deepmimic}. This strategy initializes the character at the start of each episode with a randomly chosen human pose from the reference motion. The location of the objects is updated every 512 simulation steps based on predictions made by the scene layout generator. The frame value network is updated every 32 updates for the motion imitator networks. The gradients for the motion imitation controller and the scene layout generator are calculated using the PPO algorithm~\cite{schulman2017proximal}.

\begin{figure}[t]
    \centering
    \includegraphics[width=0.45\textwidth]{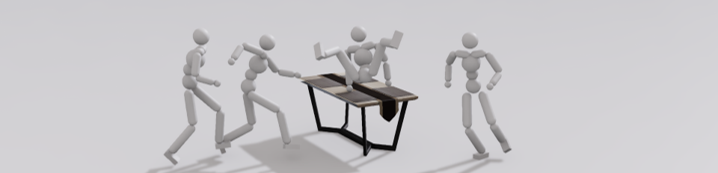}
    \caption{Illustration of the generated scene for the vaulting motion. INFERACT can be applied to a wide range of motions.}
    \label{fig:other_motions}
\end{figure}

\begin{table}%
\caption{Tracking scores and success rates on the SAMP dataset. Our method outperforms others.}
\label{tab:one}
\begin{minipage}{\columnwidth}
\begin{center}
\begin{tabular}{l clcl}
              \toprule
              \textbf{Method} & \textbf{tracking score} & \textbf{success rate} \\
              \midrule
              SUMMON & - & \quad\quad 0.53 \\
              SUMMON in physics & 0.61 &\quad\quad 0.68 \\
              Ours w/ POSA & 0.67 & \quad\quad 0.84 \\
              \textbf{Ours} & \textbf{0.67} & \quad\quad \textbf{0.84}  \\
              \bottomrule
            \end{tabular}
\end{center}
\end{minipage}
\end{table}%

\subsection{Generated Scene Layouts}
\paragraph{Comparison}
We quantitatively compare the physical plausibility of our results with those from the state-of-the-art human mesh-based scene synthesis approach SUMMON~\cite{ye2022scene} and its variant SUMMON in physics, which combines SUMMON synthesized scenes with a physics-based motion imitation controller. We define two metrics to assess the physical plausibility of the generated scenes for physics-based approaches: the tracking score and the success rate. The tracking score is computed using the tracking reward formulated in Eq.~\ref{eq:tracking_rew}. To determine a successful trial, we consider a maximum tracking error of the key body parts (hands, feet, head, pelvis) that is less than 0.3m. We also extend the definition of success rate to include the kinematics-based approach SUMMON. For this approach, a trial is considered successful if there is no significant penetration, which is determined by the non-collision score~\cite{zhang2020generating} being less than 0.85.

The visualizations of the generated results for comparison are illustrated in Fig.~\ref{fig:comparison}. According to the figure, our approach successfully generates physically plausible scenes for three different settings: a human interacting with multiple objects, a human performing complicated chair motions, and a human interacting with an object without obvious interaction affordances. In contrast, the compared method SUMMON exhibits artifacts such as severe penetration or floating on the interacting object. Furthermore, we include another strong baseline, MIME~\cite{yi2023mime}, in our qualitative comparison for scene layout generation. Similar to SUMMON, MIME also exhibits limitations in creating physically plausible scenes for human interaction, as shown in Fig.~\ref{fig:comparison_mime}. The comparison clearly demonstrates the strength of our method in producing physically plausible scenes with animated humans.

\paragraph{Diversity}
The diversity of the selected objects is demonstrated in Fig.~\ref{fig:object_selection}. As shown in the visualization results, the scene layout generator is capable of generating a scene distribution. The chair in the left figure is the most suitable for this motion and is therefore picked with the highest chance of 0.82. The selection probability decreases if the heights of the chairs do not match, as shown in the second and third examples. If the object does not have a proper affordance to support the human, like the table illustrated in the rightmost example, it has nearly zero probability of being selected. These results demonstrate the ability of our method to generate diverse object selections with plausible placements for the motion.

\paragraph{Object selection}
We further investigate whether our method can select appropriate interacting objects by varying the human motion. In this experiment, we utilize motions from SAMP in different sitting heights. From the visualizations depicted in Fig.~\ref{fig:objects}, where each example showcases the most probable selection, our method successfully chooses the suitable object for different motions.
\paragraph{Generalize to outdoor motions}
In this experiment, we test our method on outdoor motions, which presents extreme challenges for methods that rely on pre-trained contact estimators using indoor activity motions. In Fig.~\ref{fig:other_motions}, the scene layout generator is still capable of generating a reasonable scene with a table serving as the obstacle that supports the human in completing the vaulting motion. 
This demonstrates the potential versatility of our method in handling various types of interactions.

\begin{figure}[t]
    \centering
    \includegraphics[width=0.9\linewidth]{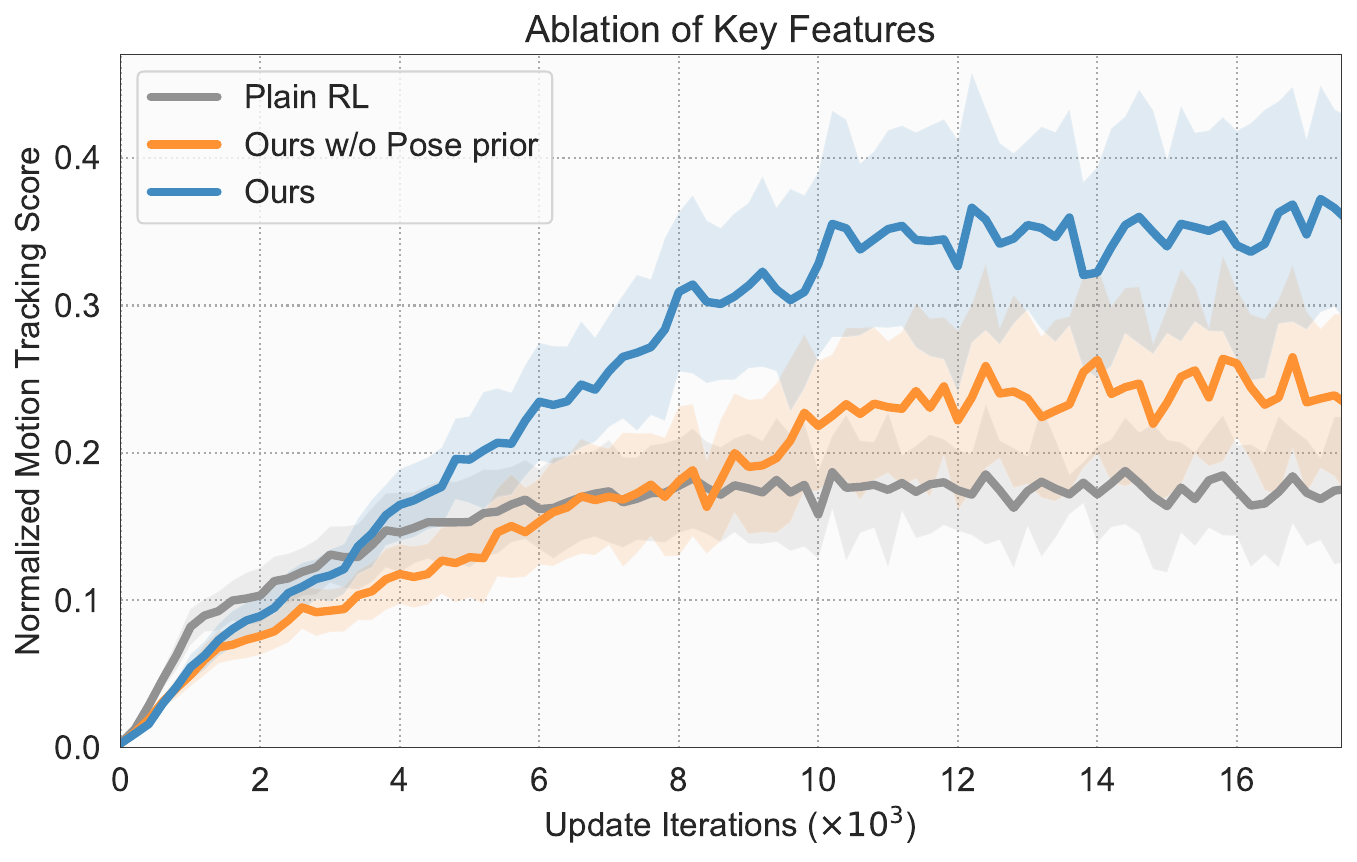}\\
    \medskip
    \includegraphics[width=0.9\linewidth]{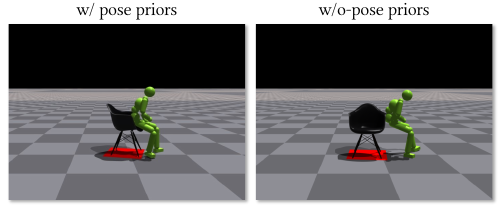}
    \caption{Results of ablation studies. The top figure depicts the learning curves of our method and ablated ones. The bottom figures illustrate the comparison between two generated scenes from ablated methods.}
    \label{fig:pose_priors}
\end{figure}

\subsection{Ablation Study}
\begin{figure}[t]
    \centering
        \includegraphics[width=0.95\linewidth]{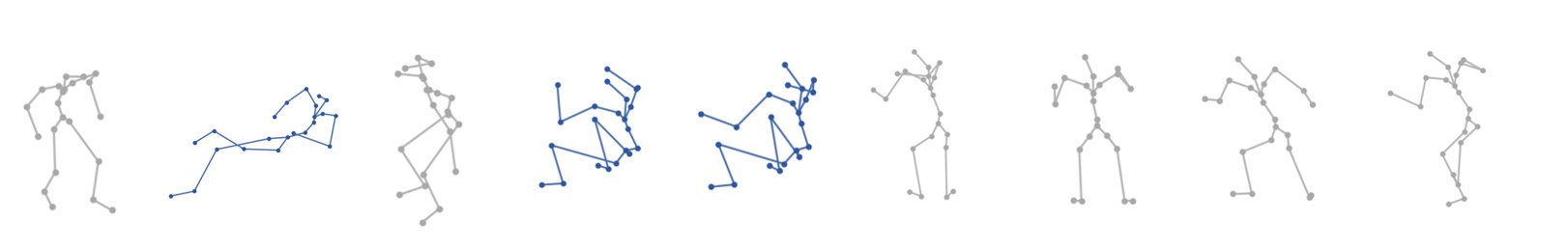}
    \medskip
        \includegraphics[width=0.95\linewidth]{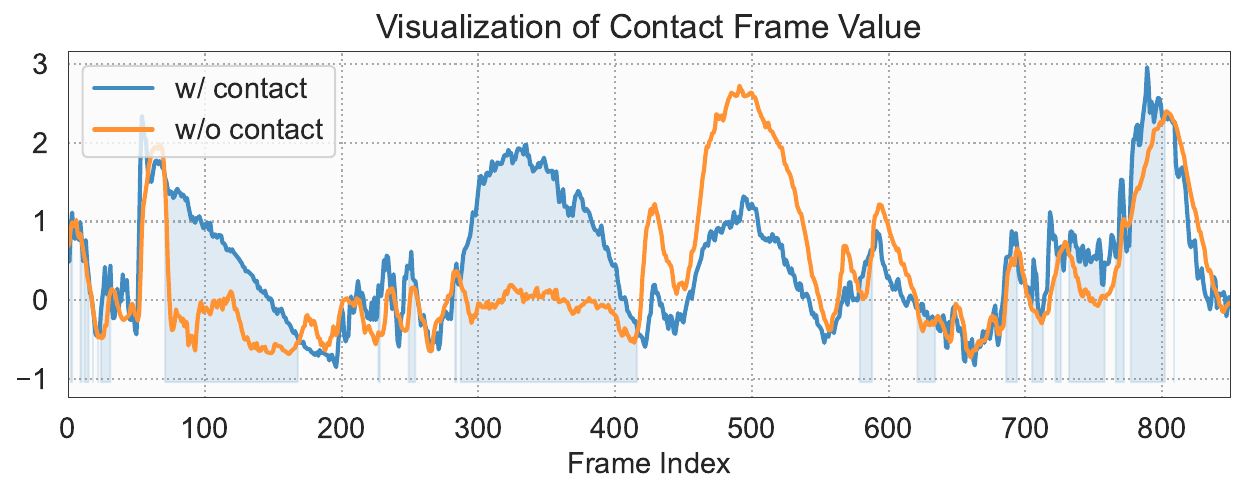}
    \caption{Visualizations of contact frame values and pseudo contact labels. The top figure presents skeleton pose snapshots, while the bottom plot displays the estimated contact frame values from the frame value networks after 500 updates. The shadow areas represent the pseudo contact labels. The visualization demonstrates the frame value network's ability to predict approximate correct contacts at an early stage of training.}
    \label{fig:contact_val}
\end{figure}

\paragraph{Reward guidance}
Firstly, we examine the significance of the reward guidance, which includes the character-object contact constraint and the pose prior guidance, on the learning efficiency of the scene layout generator. To assess the learning efficiency, We use the learning curve as a proxy to reveal the motion tracking score with respect to the update iteration. As shown in Fig.~\ref{fig:pose_priors}, we can observe a significant improvement in the methods that incorporate scene-character contact constraints compared to plain RL optimization without the scene-character contact constraint and the pose prior guidance. Regarding the pose prior guidance, the method with this feature experiences further improvement in its learning procedure and achieves a higher tracking score after the convergence point. During the experiments, we also observe additional benefits of employing the pose prior guidance. In some cases, the agent ``hacks'' the motion tracking task by learning a cheating object placement. As shown in Fig.~\ref{fig:pose_priors}, the character learns to lean against the armrest of the chair instead of sitting on the chair. With the pose prior guidance, our method effectively mitigates this issue and generates more reasonable results.
\paragraph{Pseudo contact labels}
We further investigate the efficacy of our proposed pseudo contact label estimation method by replacing it with ground truth contacts, in providing pose prior guidance. This modified approach, referred to as ours with POSA, utilizes the pre-trained POSA model to predict the contact frames. As the results presented in Table.~\ref{tab:one}, our method maintains comparable performance even without prior knowledge of motion contacts. Moreover, our method can be directly applied to novel motions without the need for training the contact estimator. We also visualize the estimated contact values for a sitting motion in Fig.~\ref{fig:contact_val}, where the estimated pseudo contact labels, indicated by the blue shaded area, are approximately consistent with the motion semantics.

\section{Conclusion}
We presented a method INFERACT that generates plausible scene layouts supporting realistic human-scene interaction in a physics-based environment. Our method introduces physical constraints and motion dynamics as constraints and optimizes both scene configurations and the simulated character controller to maximize the motion tracking objective using reinforcement learning. The generated scenes for the motions from SAMP and PROX show enhanced physical plausibility compared with the state-of-the-art human mesh-based scene synthesis method. 
The results of INFERACT also demonstrate the ability to generate versatile scenes for a single motion and select the most matching objects for different motions. 

\paragraph{Limitations and future work.} INFERACT still has certain limitations. One notable limitation is that it approximates all physical interactions using rigid body contact. However, this approach may not accurately capture the complex dynamics of human-scene interaction in the real world. Consequently, visual artifacts, such as motion jitters during contact, may still exist. Another limitation arises from the restricted range of supported interaction relationships. To expand the scope of human-scene interactions, a potential future direction is to explore a more comprehensive optimization objective that encompasses a wider range of interaction categories.

\begin{acks}
We thank Lei Han and He Zhang for their insightful and stimulating discussion on the concept. This work was supported by Tencent and Hong Kong Innovation and Technology Commission (ITS/307/20FP).
\end{acks}

\bibliographystyle{ACM-Reference-Format}
\bibliography{reference}

\appendix
\section{Details of State Representation}
The local proprioceptive observation $s^H_t$ includes various features that describe the states of the animated humanoid character. These features include the height and orientation of the root, the velocity of the root, the angular velocity of the root, joint rotations represented using 6D normal-tangent vectors, joint velocities, and the relative positions of key body parts (such as hands and feet) with respect to the root, referred as to \textit{keypos}. All of these features are calculated in the local coordinate system of the humanoid root. The scene layout state $s^E_t$ incorporates two components. Firstly, it includes an index indicating the selected object from the object set $O$. Secondly, it includes a 3D vector that indicates the placement of the object. This vector consists of the global translations on the $X-Y$ plane and the rotation angle in the $Z$ axis. The relative tracking feature $s^R_t$ consists of the features of the reference human pose. It includes the local root position, orientation, and linear and angular velocities of the reference pose in the coordinate of the animated human root. Additionally, it includes joint rotations and velocities, and \textit{keypos} of the reference human pose.

\section{Computation of Tracking Rewards}
The tracking reward for the character's root position, root orientation, and root velocity are calculated as follows:
\begin{displaymath}
    r^p_t = \exp{(-10 \cdot \|p_t - \hat{p_t}\|)},
\end{displaymath}
\begin{displaymath}
    r^o_t = \exp{(-5 \cdot \|o_t - \hat{o_t}\|)},
\end{displaymath}
\begin{displaymath}
    r^v_t = \exp{(-1 \cdot \|\dot{p}_t - \hat{\dot{p}}_t\|)},
\end{displaymath}
where $p, o, \dot{p}$ denotes the root position, root orientation, and root velocity of the character, and $\hat{\cdot}$ denotes the desired value.
The reward terms for the joint positions and velocities are formulated to:
\begin{displaymath}
    r^{jp}_t = \exp{\Big(-2 \cdot \sum_{j=1}^{N} (q^j_t - \hat{q^j_t})^2\Big)}
\end{displaymath}
\begin{displaymath}
    r^{jv}_t = \exp{\Big(-0.1 \cdot \sum_{j=1}^{N} (\dot{q}^j_t - \hat{\dot{q}}^j_t)^2\Big)},
\end{displaymath}
where $q^j$ and $\dot{q}^j$ represent the joint position and velocity for each DoF, and $N$ is the number of DoFs in the humanoid character.
The reward for key body positions is
\begin{displaymath}
    r^{k}_t = \exp{\Big(-10 \cdot \frac{1}{K} \sum_{k=1}^{K} \|p^k_t - \hat{p^k}_t\|\Big)},
\end{displaymath}
where $p^k$ denotes the position of the $k^{\rm th}$ body.
The weights for reward components are $w^p=0.2, w^o=0.05, w^v=0.05, w^{jp}=0.45, w^{jv}=0.05, w^k=0.15$.

\end{document}